\documentclass[runningheads]{llncs}
\usepackage{graphicx}
\usepackage{amsmath}
\usepackage{caption}
\usepackage{subcaption}
\DeclareMathOperator{\sgn}{Sgn}
\usepackage{amssymb}
\usepackage{amsmath}

\begin{document}
\title{Strategising template-guided needle placement for MR-targeted prostate biopsy}
\author{Iani JMB Gayo, Shaheer U. Saeed, Dean C. Barratt, \newline
Matthew J. Clarkson, Yipeng Hu}
\authorrunning{I. Gayo et al.}

\titlerunning{Strategising needle placement for MR-targeted prostate biopsy}

\institute{Department of Medical Physics and Biomedical Engineering, \and
Wellcome/EPSRC Centre for Interventional and Surgical Sciences, \and
Centre for Medical Image Computing,
\\ University College London, UK }
\maketitle              % typeset the header of the contribution
\begin{abstract}
Clinically significant prostate cancer has a better chance to be sampled during ultrasound-guided biopsy procedures, if suspected lesions found in pre-operative magnetic resonance (MR) images are used as targets. However, the diagnostic accuracy of the biopsy procedure is limited by the operator-dependent skills and experience in sampling the targets, a sequential decision making process that involves navigating an ultrasound probe and placing a series of sampling needles for potentially multiple targets. This work aims to learn a reinforcement learning (RL) policy that optimises the actions of continuous positioning of 2D ultrasound views and biopsy needles with respect to a guiding template, such that the MR targets can be sampled efficiently and sufficiently. We first formulate the task as a Markov decision process (MDP) and construct an environment that allows the targeting actions to be performed virtually for individual patients, based on their anatomy and lesions derived from MR images. A patient-specific policy can thus be optimised, before each biopsy procedure, by rewarding positive sampling in the MDP environment.
Experiment results from fifty four prostate cancer patients show that the proposed RL-learned policies obtained a mean hit rate of 93\% and an average cancer core length of 11 mm, which compared favourably to two alternative baseline strategies designed by humans, without hand-engineered rewards that directly maximise these clinically relevant metrics. Perhaps more interestingly, it is found that the RL agents learned strategies that were adaptive to the lesion size, where spread of the needles was prioritised for smaller lesions. Such a strategy has not been previously reported or commonly adopted in clinical practice, but led to an overall superior targeting performance, achieving higher hit rates (93\% vs 76\%) and measured cancer core lengths (11.0mm vs 9.8mm) when compared with intuitively designed strategies.

\end{abstract}
\section{Introduction}
%Prostate Cancer (PC) is the predominant cancer in men in the Western World \cite{pc_stat}. Biopsy is used to diagnose PC, where tissue samples are collected to establish evidence of malignant disease. Ultrasound imaging (ultrasound) is commonly used to guide biopsy needles into position, however many lesions are isoechoic and not clearly visible on ultrasound scans \cite{us_prostate}. 
Recent development in multiparametric MR imaging (mpMRI) techniques provides a means of noninasive localisation of suspected prostate cancer \cite{mpMRI}, which enables clinicians to target these lesions during the follow-up ultrasound-guided biopsy for further histopathology confirmation. This MR-targeted approach has been shown to reduce both the false positive and false negative detection, compared to previously adopted random biopsy \cite{mpMRI,picture_study}, and subsequently motivated research in developing multimodal MR-to-ultrasound image registration \cite{registration_error}.

%By efficiently sampling clinically significant cancers identified during planning, this targeted approach potentially reduces the chances of false positive and negatives in detecting cancers, compared to previously adopted random biopsy. 

%increases the detection rate of of early-stage cancer, by efficiently sampling clinically significant cancers identified during planning.

%This can be used for planning, to enable clinicians to target these lesions during biopsy, allowing for precise sampling for subsequent histopathology examination. This targeted biopsy approach potentially increases the detection rate of of early-stage cancer, by efficiently sampling clinically significant cancers identified from the planning stage. 

Needle sampling of the MR-identified targets, with or without registration errors, can still be a challenging and arguably overlooked task. Operator expertise was found to be an important predictor in detecting clinically significant prostate cancer \cite{interoperator_variance}. Planning strategies is important for navigating the ultrasound probe, to better observe the targets with respect to imaging, and for manual needle positioning. In transperineal biopsy, the introduction of brachytherapy templates (See Fig.1 for an example) helps the needle deployment - a procedure that is of interest in this study, but choice between $13 \times 13$ grid positions remains a subjective decision. For example, a common clinical practice aims at the target centre, but it has been shown to yield an insufficient sampling of the heterogeneous cancer~\cite{heterogeneity} and possibly an inferior diagnostic accuracy in terms of disease-representative grading \cite{aim_for_centre}, compared with more spread needle placement. The design of an optimum strategy is further complicated by the need of multiple needles for individual targets, for maximising the hit rate, and the multifoci nature of prostate cancer, which requires repeated sampling of one or more targets. 

To the best of our knowledge, there has not been any computer-assisted sampling strategy that takes into account the previous needle deployment(s) or quantitatively optimises patient-and-target-specific needle distribution. In summary, improving the targeting strategy may help reduce the significant false negative rate found in MR-targeted biopsy (reported being as high as 13\% \cite{smart_target}), and hence improves the chance of early cancer detection for patients.

Reinforcement learning (RL) has been proposed for medical image analysis tasks \cite{rl_medimg}, such as landmark detection \cite{landmark}, plane finding \cite{plane_finding}, and for surgical planning such as hysterectomy \cite{hysterectomy} and orthopaedic operations \cite{orthopaedic}. It has also been used for needle path planning in minimally-invasive robotic surgery \cite{needle_planning}, \cite{needle_planning_2}. It is its ability to learn intelligent policies for sequential decision making that provides a potential solution to problems without requiring direct supervision for each action taken, a common constraint in developing machine learning-assisted methods for complex and skill-demanding surgical and intervention applications. This makes RL suitable for finding an optimal targeting strategy, which requires complex decisions for which there is no established best method. 
%Simplified 2D environments were constructed to represent the procedures, with limited patient data sets which may not represent the complexities of the procedures. More recently \cite{orthopaedic}, proposed an orthopedic surgical planning framework where clinicians marked the learned plans as better than gold standard.

In this study, we investigate the feasibility of using RL to plan patient-specific needle sampling strategies, optimised in pre-operative MR-derived RL environments. We present experimental results based on clinical data from prostate cancer patients and compare the proposed method, using a set of clinically important metrics, to baseline strategies that are designed by human intuition and an interactive targeting performed by two observers. %Such learned strategies could help clinicians sample and detect more clinically significant cancers, and obtain accurate diagnosis that allow patients to receive appropriate follow-up care. 
We conclude by reporting a set of interesting observations that demonstrate the benefit of using the proposed RL-learned patient-specific strategies. These indeed adapted effectively to individual procedures and varying targets, for improved final performance of the sequential target sampling. Consistent hit rates were achieved with less variance for both smaller and larger lesions as a result of the learned adaptive strategies.

\section{Method}

%%%%%%%%%%%%%%%%%%%%%%%%%%%%%%%%%%%%%%%%%%%%%%%%%%
% SS suggested

%\subsection{Reinforcement learning formalisation}
%Reinforcement learning (RL) is based on interactions between an agent and an environment. 
The agent-environment interactions are modelled as a Markov decision process (MDP), and summarised 
as a 4-tuple $\langle \mathcal{S}, \mathcal{A}, r, p \rangle$, where $\mathcal{S}$ and $\mathcal{A}$ are the state and action spaces consisting of all possible observed states as input and actions as output for the agent, respectively. $r: \mathcal{S} \times \mathcal{A} \rightarrow \mathbb{R}$ is the reward function which maps state-action pairs to a real value. The state transition distribution is defined by $p: \mathcal{S} \times \mathcal{S} \times \mathcal{A} \rightarrow [0, 1]$ which denotes probability of transitioning to the next state, given the current state-action pair. In this section, we develop an environment for template-guided biopsy sampling of the cancer targets, the MDP components and the policy learning strategy.

%The agent-environment interactions can be formulated using these elements. We describe the environment and the relevant components in the following subsections.

%A system of agent-environment interactions may be constructed with these elements. We describe the environment and the relevant components in the following subsections.

%%%%%%%%%%%%%%%%%%%%%%%%%%%%%%%%%%%%%%%%%%%%%%%%%%

\subsection{Patient-specific prostate MR-derived biopsy environment}
\label{sec:method.env}
\vspace{-5mm}
\begin{figure}[ht!]
     %\centering
     \begin{subfigure}[b]{0.4\textwidth}
         %\centering
         \includegraphics[width=\textwidth]{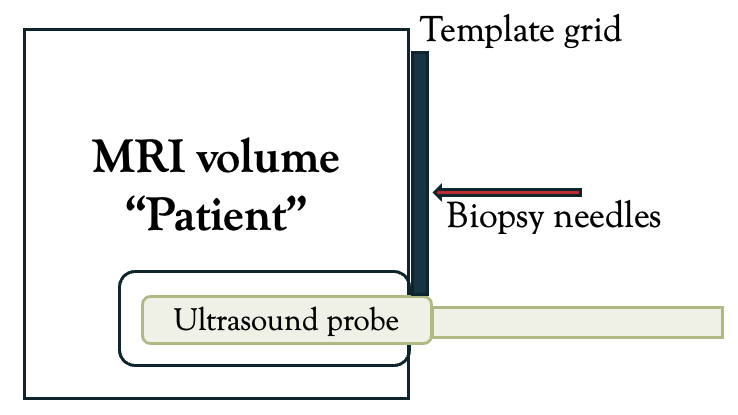}
         \caption{Environment}
         \label{fig:simulation}
     \end{subfigure}
     %\hfill
     \begin{subfigure}[b]{0.25\textwidth}
         \centering
         \includegraphics[width=\textwidth]{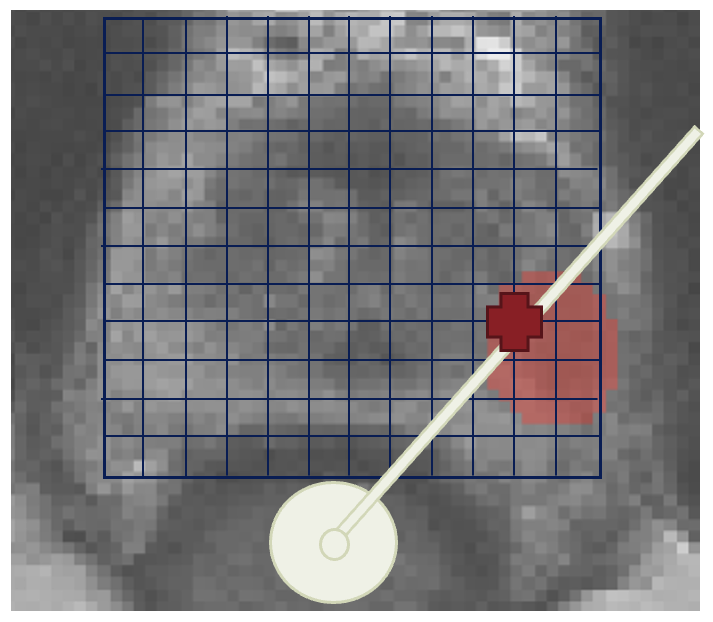}
         \caption{Template grid}
         \label{fig:template_grid}
     \end{subfigure}
     \hfill
      \begin{subfigure}[b]{0.3\textwidth}
         \centering
         \includegraphics[width=\textwidth]{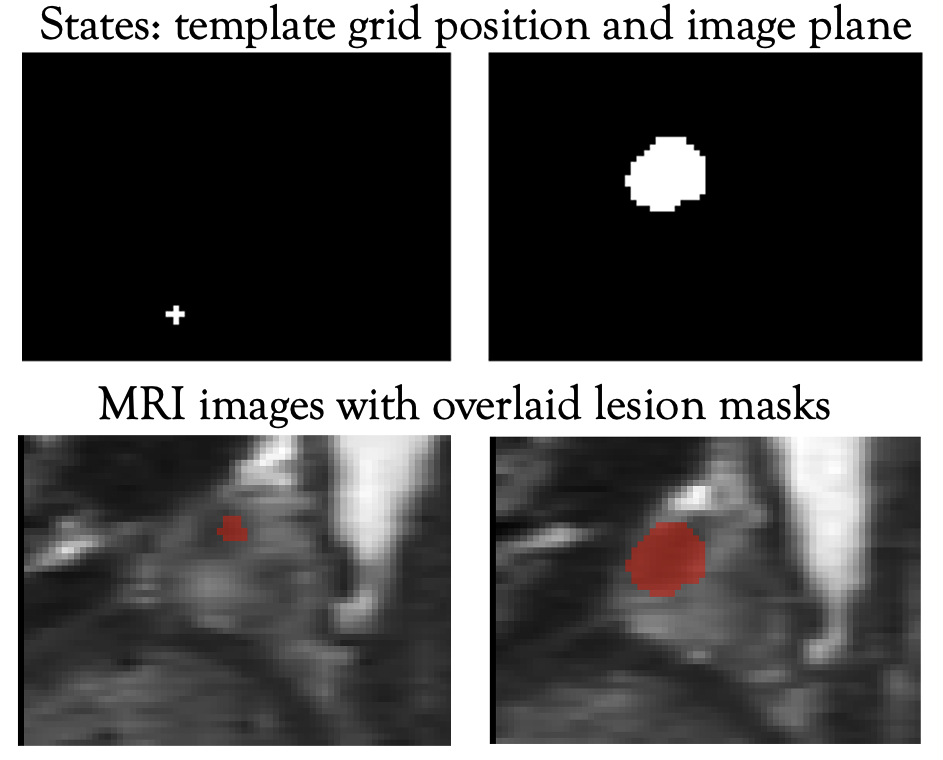}
         \caption{States and targets}
         \label{fig:obs}
     \end{subfigure}
        \caption{Simulated biopsy procedure environment. (a) Placement of ultrasound probe and template grid within the MRI volume. (b) Visualisation of ultrasound probe rotation which is always aligned with the chosen template grid position. (c) Examples of (top) states and (bottom) overlaid MR-identified targets}
        \label{fig:environment_vis}
\end{figure}
\vspace{-5mm}

%During biopsy, the needle can be inserted directly through the rectum, known as a transrectal (TR) ultrasound biopsy. An alternative approach, transperineal (TP) is to insert the needles through the perineum, often using a template grid known as a brachytherapy grid, which is also used for targeted treatments. In this work, the TP approach is simulated as the same template grid can be used for targeted treatment procedures \cite{}
%Two biopsy approaches exist for prostate biopsy : transrectal and transperineal. The differences between the two approaches can be visualised in figure \cite{}. In TP, needles are inserted through the perineum, often using a template grid known as a brachytherapy grid, which is also used for targeted treatments. 

The environment is illustrated in Figure \ref{fig:simulation} for the targeted biopsy procedures, where virtual biopsy needles are inserted through the perineum via a brachytherapy template grid consisting of 13x13 holes that are 5mm apart. Other needle-based treatments such as cryotherapy, brachytherapy and radiofrequency ablation \cite{focal_therapy} may also be applicable but are not discussed further in this paper. The position of the transrectal ultrasound probe is approximated within the rectum directly underneath the prostate gland, with a fixed distance to the template grid such that the top of the probe is aligned with the lower side of the template, as shown in the Figure \ref{fig:simulation}. Both anatomical and pathological structures can be sampled, at any sagittal ultrasound imaging plane given an arbitrary angle, as illustrated in Figure \ref{fig:template_grid}. %, \ref{fig:obs}.

We summarised a number of considerations in designing and constructing the adopted biopsy environment as follows. 1) The prostate gland from each MR volume, the MR-identified targets (as in Figure \ref{fig:obs}), and key landmarks such as position of the rectum are all manually segmented from individual patients to construct the biopsy environment. Automated methods for segmenting these regions of interest have been proposed, e.g. \cite{prostate_segmentation}, \cite{tumour_segmentation}. 2) Binary segmentation are provided as observations for the RL agents, as opposed to ultrasound image intensities, which are neither available during planning nor straightforward to synthesize from MR images. We argue that the use of binary representation would be more robust to train the RL agents and the resulting methods are more likely to generalise to different procedures and planning MR images, especially given the existing MR and ultrasound segmentation and registration algorithms described above. 3) Uncertainty in MR-to-ultrasound registration can and should be added to the segmented regions, together with other potential errors in localising these regions during observation such as observer variability in manual segmentation used in this work. We would like to point out that, however, these localisation errors are unlikely to be independent to each other and the dependency of RL model generalisability on how precisely these need to be modelled remain open research questions. Sec.~\ref{sec:exp} discusses further details adopted in our experiments. 4) In the presented experiments, we focus on targeting the index lesions, those are of largest volumes in each case, to provide first results that show the efficacy of modelling the dynamic biopsy sampling process. However, the described MDP should be directly applicable for and likely to be more effective in cases with multiple lesions.

\subsection{The MDP components}

%A reinforcement learning problem can be framed as a markov decision process, for which the state of the environment, the actions of the agent, the reward function can be defined. 

%\begin{equation}
%    MDP =  < S, A, R, P, \gamma >
%\end{equation}

%%%%%%%%%%%%%%%%%%%%%%%%%%%%%%%%%%%%%%%%%%%

% SS suggested
\paragraph{State} - At a given time point $t$ during the procedure, the agent receives information about its current state $s_t\in\mathcal{S}$ : the chosen grid point and the re-sampled 2D image plane obtained by rotating the probe to the current template grid position, as in Figure \ref{fig:obs}. This current position is determined by the previous action. This is to test the scenario with least assumptions, where the overall 3D anatomical and pathological information may be corrupted or unreliable due to intra-procedural uncertainties from patient movement and outdated registration. 

%as if the probe is aligned and rotated at the chosen template grid position.as if the probe is aligned and rotated to the chosen template grid position. All three components are of the same size and are stacked to form a single observed state at time-step $t$, $s_t\in\mathcal{S}$.

%The images are obtained as if the probe is aligned and rotated at the chosen template grid position.
%%%%%%%%%%%%%%%%%%%%%%%%%%%%%%%%%%%%%%%%%%%

%\subsubsection{Actions (A)} - The agent can take relative actions $x_{\delta}$ and $y_{\delta}$ from its current position on the template grid. The range delta can take is from (-15,15). To obtain a robust plan, the agent is initialised on different starting positions within the centre of the template grid, and is given the task to find its way towards the lesion. 
%%%%%%%%%%%%%%%%%%%%%%%%%%%%%%%%%%%%%%%%%%%
% SS suggested

\vspace{-6pt}
\paragraph{Actions} - The agent takes actions $a_t\in\mathcal{A}$ which modify its position on the template grid. These actions are relative to the current position of the agent $(i, j)$ and are defined as $a_t=(\delta_i, \delta_j)$ such that the new position is given by $(i+\delta_i, j+\delta_j)$, where $\delta_i, \delta_j\in[-15,15]$. By formulating this relative grid-moving action, we consider the biopsy needle is positioned on the image plane, with an insertion depth that overlaps the needle centre and the centre of the observed 2D target, subject to small predefined positioning errors in each direction. These are commonly adopted practice though not strictly enforced, and we found that increasing the flexibility by independently positioning the ultrasound probe and needle may unnecessarily make the training difficult to converge.

\vspace{-6pt}
\paragraph{Reward} - The reward at the time $t$ is computed based on the reward function $R_t = r(s_t, a_t)$, during training. The agent is rewarded positively if the fired needle obtains samples of the lesion. A penalty is given when chosen grid positions are outside of the prostate, to avoid hitting surrounding critical structures. From initial experiments, it was found that a greater reward of +5 lead to a faster convergence during training, encouraging the agent to hit the lesions, whilst a penalty of -1 was enough to deter the agent from firing outside the prostate gland. Reward shaping is also introduced to guide the agent towards the lesion, thereby speeding up the learning process. Similar to \cite{landmark} and \cite{plane_finding}, a sign function $\sgn$ of the difference between $dist_{t-1}$ and $dist_{t}$ is computed, where $dist_{t}$ represents Euclidean distance between target centre and needle trajectory at time $t$.

\vspace{-5mm}
\begin{equation}
\label{eqn:nr}
r = 
   \begin{cases}
$+5$ & \textit{if biopsy needle intersected with target} \\
$-1$ & \textit{if biopsy needle placed outside prostate} \\
\sgn(dist_{t-1} - dist_{t})   & \textit{otherwise}
\end{cases} 
\end{equation}

\subsection{Policy learning}

% At time-step $t$, the state transition may then be described by $p(s_{t+1} | s_t, a_t)$ which denotes probability of next state $s_{t+1}\in\mathcal{S}$ given current state $s_t\in\mathcal{S}$ and current action $a_t\in\mathcal{A}$. The goal is to learn parameters for a parametric policy from which an action is sampled $a_t\sim\pi_\theta(\cdot | s_t)$ which represents the probability of performing action $a_t$ given state $s_t$.  Following the policy $\pi_\theta$ and the transition distribution $p$, leads to a trajectory $(s_1, a_1, R_1, \cdots, s_T, a_T, R_T)$. The accumulated reward, when following parametric policy $\pi_\theta$, is $Q^{\pi_\theta} (s_t, a_t) = \sum_{k=0}^{T} \gamma^k R_{t+k}$. $\gamma$ is a discount factor set to 0.9. The optimal parameters can be found $\theta^* = \text{argmax}_{\theta} \mathbb{E}_{\pi_\theta}[Q^{\pi_\theta}(s_t, a_t)]$. The optimal policy is thus defined as $\pi_{\theta^*}$.

The navigation and sampling strategy is parameterised by a policy neural network $\pi_\theta$, with parameters $\theta$, quantifying the probability of performing action $a_t$ given state $s_t$. Agent's actions then can be sampled from the policy, $a_t\sim\pi_\theta(\cdot | s_t)$. During the policy training, the accumulated reward $Q^{\pi_\theta} (s_t, a_t) = \sum_{k=0}^{T} \gamma^k R_{t+k}$ is maximised, where $\gamma$ is a discount factor set to 0.9, to obtain the optimal policy $\pi_{\theta^*}$, $\theta^* = \text{arg max}_{\theta} \mathbb{E}_{\pi_\theta}[Q^{\pi_\theta}(s_t, a_t)]$. %The optimal policy is thus defined as $\pi_{\theta^*}$., when following parametric policy $\pi_\theta$, known as
With continuous actions, policy gradient (PG) and actor-critic (AC) algorithms can thus be adopted for the optimisation.

%AC methods incorporate a second network, known as the critic which estimates the value-function $Q^{\pi_\theta} (s_t, a_t)$. The policy network, also known as the actor, is then optimised in the direction suggested by the critic \cite{spinning_up_ai}. 

%The PPO algorithm \cite{PPO} is used in our problem as this algorithm guarantees a monotonic reward improvement and stability of training \cite{ppo_paper}. The policy network consists of a CNN based on a ResNet18 architecture, followed by linear layers to estimate the action. Adam optimiser is used, with a learning rate of 0.0001. 

%An entropy coefficient of 0.001 is included to encourage exploration. 
%%%%%%%%%%%%%%%%%%%%%%%%%%%%%%%%%%%%%%%%%%%

\section{Experiments}
\label{sec:exp}

\paragraph{Data set} - The T2-weighted MR images and their segmentation were acquired from 54 prostate cancer patients. These were obtained as part of clinical trials, PROMIS \cite{mpMRI} and SmartTarget \cite{smart_target}, where patients underwent ultrasound-guided minimally invasive needle biopsies and focal therapy procedures.

\vspace{-6pt}
\paragraph{RL algorithm implementation} - An agent was trained for each patient individually using the Stable Baselines implementation of PPO \cite{Stablebaselines}. Each agent was trained for 120,000 episodes and a model was selected with the highest average reward after 10 episodes. Each episode was limited to a maximum of 15 time steps, but can terminate early if any 5 needles hit the lesion. At each episode the agent is initialised at random starting positions on the template grid. %During evaluation, different starting positions are used that were not included during training. 
Based on estimated registration error reported previously \cite{registration_error}, random localisation error was added in the observed states, equivalent to a Gaussian noise with a standard deviation of 1.73mm in each of the x, y and z coordinates, or a mean distance error of 3mm. 
The PPO algorithm \cite{ppo} was used in reported results, as it guarantees a monotonic reward improvement and stability of training. However, we also report a lack of substantial difference in performance to other tested algorithms, including DDPG \cite{DDPG} and SAC \cite{SAC}. The policy network was based on ResNet18 \cite{resnet} architecture, with an additional fully-connected layer for a linear output. An Adam optimiser was used, with a learning rate of 0.0001. It could be of future interest to compare further network architectures and PG/AC algorithms on the proposed RL problem, but is considered beyond the scope of current work.
%followed by linear layers to estimate the action.
%\subsubsection{Prostate biopsy metrics} -
\vspace{-6pt}
\paragraph{Biopsy performance metrics} - To quantitatively assess target sampling performance, three biopsy-specific metrics are used in this study: hit rate (HR), cancer core length (CCL) and needle area (NA). The HR is the number of needle samples that contains the target, i.e. positive samples, divided by the number of needles fired. Five needles are chosen to represent the maximum number typically used in targeted biopsy \cite{needles_biopsy}. CCL is the total length (in mm) overlapping the target, i.e. the sampled target tissue, in individual needles. CCL $>=6mm$ often indicates clinical significance \cite{clinically_significant}. NA estimates the coverage of all fired needles in each episode, defined as the area of an approximating ellipse, $NA = \pi * std_x \times std_y$, where $std_x$ and $std_y$ are standard deviations in the needle navigating x-y plane, defined by the template grid position.  %These describe needle efficiency and coverage.

%These describe the needle efficiency and coverage, which are two factors that we aim to optimise in this framework. 

%metrics that are specific to prostate biopsy are used These metrics can be used to quantitatively describe the learned navigation strategy. 

%\begin{equation}
%   $    HR = \frac{\# Needle_{hit}}{ \# Needle_{fired}}
%$$\end{equation}

%intersection between the needle sample and the lesion in mm. This indicates how much lesion the sampled needle contains.

    % \begin{equation}
    %     Area_{needles} = \pi * stdev_x * stdev_y
    % \end{equation}
    
\vspace{-6pt}
\paragraph{Baseline strategies} - Two strategies were compared with the proposed agent, designed to provide an estimate of what clinicians are likely to achieve in practice. For a fair comparison, the same observed targets, the states, and starting positions were used. Student's t-tests were used when comparison is made at a significance level of $\alpha=0.05$, unless otherwise indicated.
%For the trained agent, the last 5 needle samples are assumed as fired positions.
The first strategy \textit{(Sweeping strategy)} adopted a simple sweeping of the biopsy needle together with the ultrasound probe, from left to right in a 5mm interval. The target was sampled at the centre of the observed target, i.e. fired, once an image plane is encountered with a lesion. The second strategy \textit{(Scouting strategy)} moves the virtual probe to scout all candidate positions that samples the target, before 5 random ones were selected as fired positions among these candidates.
Inter- and intra-operator variance is one of the most important factors that impact the performance of a sampling strategy. For the baseline strategies, additional Gaussian noise was added to the chosen needle/probe positioning with a varying bias and a varying standard deviation (SD). Increasing the SD leads to higher uncertainty in placing the needles, while the bias indicates a targeting strategy that does not aim for the target centre, e.g. for avoiding empty cores or urethra. The experiments were repeated using different combinations of the two variables (each ranging between 0 to 10mm), to test different strategies.

% \textbf{Patient-specific sampling plans} The performance of learned strategies is evaluated for individual patients with varying lesion sizes. We compare the learned biopsy metrics for smaller and larger lesions to observe differences in the policies and evaluate the agents' ability to adapt its policies for different patients. 

%Varying bias and SD values represents multiple clinicians which choose the fired positions from different sampling distributions. These strategies could potentially be used as labels for a supervised model where observed planes can be used to estimate the best needle firing positions that target the lesion centres. 

\vspace{-6pt}
\paragraph{Interactive experiments by two observers} - Two human operators, one computer scientist and one biomedical imaging researcher, interacted with a custom-made interface that displays the current template grid position and image plane observed. They were asked to choose where to sample and how far to spread the needles. This is a simple interactive experiment to provide preliminary results that can be compared with the other described strategies.

\section{Results}
\paragraph{Learned strategy performance} - From Table \ref{tab:simple_policy}, the agent outperforms both baseline strategies in HR and CCL (both p-values$<$0.005), but not in NA, with noticeably smaller variability in HR. Different levels of bias did not lead to significantly different targeting results, while higher SD increased the spread of needles, but reduced CCL and HR. In general, the results between the sweeping and scouting strategies were not found statistically different in CCL and NA. The scouting strategy resulted in an increased HR (p-value$<$0.05) compared to sweeping, but does not outperform the agent which still obtains the highest HR. %The agent outperforms both baseline strategies and hits the lesions more accurately, by reducing the spread of the needles fired. %, which suggests the saved positions helped . %In summary, the agent learned the optimal positions, more robust to the included localisation errors described in Sec.\ref{sec:exp}. 

\vspace{-5mm}
\begin{table}
\caption{Summary of biopsy performance from the RL agent (top row) and the sweeping and scouting strategies for different bias and SD combinations}
\label{tab:simple_policy}
%\centering
\begin{tabular}{|cc|c|c|c||c|c|c|}
\hline
\multicolumn{2}{|c|}{\bfseries } & \multicolumn{3}{|c||}{\bfseries Baseline 1 (Sweeping) } & \multicolumn{3}{|c|}{\bfseries Baseline 2 (Scouting)}\\
\hline
Bias & SD & CCL(mm) & HR(\%) & NA($mm^2$) & CCL(mm) & HR(\%) & NA($mm^2$) \\
\hline
\hline 
\multicolumn{2}{|c|}{\bfseries Agent} &  11.13±3.43 & 93.40 ± 11.44 &
22.14±18.18 & 11.13±3.43 & 93.40 ± 11.44 &
22.14±18.18\\
\hline
\hline

%bias 0 
\hline
 0 & 0 & 7.95±3.00
 & 53.36±35.59
& 23.67±16.57 & 8.40±2.65
 & 61.10±30.01
& 31.31±28.53
 \\

 \hline
 0 & 5 & 7.45±3.19
 & 49.43±34.10 
& 56.81±50.25 

&
8.32±3.04 & 55.19±30.46 & 39.84±20.48\\

\hline
 0 & 10
& 4.89±3.29
& 30.94±28.71 
 & 108.00±86.24 & 5.67±4.32
& 43.70±31.99 
 & 114.35±71.20\\
 \hline
 \hline
 
%bias 5
5 & 0
& 8.90±4.00
& 53.69±28.44
 & 36.53±28.91
 
 & 8.32±3.00
& 65.19±30.04
 & 36.53±28.91

\\
\hline
5 & 5 &
7.88±3.52 
& 51.32±32.94
& 60.44±47.62

&
7.28±3.17 &
54.48±31.04 &
72.08±40.89\\
\hline
5 & 10 &
6.06±3.95
& 41.51±31.58
& 86.15±63.15

&
5.74±4.41
& 44.82±31.02
& 113.88±79.60

\\
\hline
\hline

%bias 10
10 & 0 &
7.29±3.08 &
47.80±34.16
&
27.47±18.38

&
7.65±3.05 &
54.81±31.13
&
27.59±22.95
\\
\hline 
10 & 5 &
7.14±3.38
&
42.44±28.00 &
52.78±48.02
&
6.40±3.64 &
50.74±29.00 &
55.45±42.36\\
\hline
10 & 10 &
4.74±3.46 &
28.70±27.22 &
106.17±115.95 

&
5.98±3.42 &
47.04±31.22 &
112.68±95.85
\\

\hline
\end{tabular}
\end{table}
\raggedbottom

\vspace{-15mm}
\begin{table}[ht!]
    \centering
    \caption{Obtained biopsy strategy metrics for the agent and two human observers}
    \begin{tabular}{|c|c|c|c|}
    \hline
         \textbf{Observers} &  \textbf{CCL (mm)} & \textbf{HR(\%)} & \textbf{NA ($mm^2$)}\\
         \hline
        Agent &  11.13 ± 3.43  & 93.40 ± 11.44 & 22.14 ± 18.18\\ 
        \hline
        Observer 1 &  9.71 ± 3.78 & 66.30 ± 20.55 & 42.85 ± 23.36 \\
        \hline
        Observer 2 & 9.83 ± 3.89 & 76.30 ± 19.50 & 64.90 ± 38.84 \\
        \hline
% & 42.85 ± 23.36
    \end{tabular}
    
    \label{tab:my_label}
\end{table}
% % %%%\vspace{-5mm}
% \begin{table}[ht!]
% \caption{Quantitative comparison of biopsy strategy learned by agent and two human operators}\label{tab:exp_human}
% \centering
% \begin{tabular}{|c|c|c|c|}
% \caption{Quantitative comparison of biopsy strategy learned by agent and two human operators}\label{tab:exp_human}
% \hline
% Operators &  CCL (mm) & HR(\%) & Coverage($mm^2$)\\
% \hline
% Agent &  11.13 ± 2.43  & 82.22 ± 27.10 &
% 22.14 ± 18.18\\ 
% \hline
% Operator 1 &  9.71 ± 3.78 & 66.30 ± 20.55
% & 42.85 ± 23.36 \\
% \hline
% Operator 2 & 9.83 ± 3.89 & 76.30 ± 19.50 & 64.90 ± 38.84 \\
% \hline
% \end{tabular}
% \end{table}
% \raggedbottom
\vspace{-5mm}

\vspace{-6pt}
\paragraph{Interactive experiments by two observers} - The agent outperforms both observers in CCL (p-values=0.020 \& 0.040), but its NA values are more than double that of the agents, suggesting a potential trade-off between sampling coverage and precision. For HR, the agent outperforms both observers (p-value $<$ 0.001), which demonstrates that the agent could achieve an overall comparable performance as human observers, with a significantly higher CCL. % which is an important diagnostic measure for biopsy.

%This further emphasises the benefits of the learned navigation strategies in optimising CCL, allowing for more accurate diagnostic measures during biopsy. 

%This however still demonstrates that the agent is able to achieve similar performance as human operators with regards to needle firing efficiency. 

%These results are reasonable, as the agent is not rewarded for spreading the fired biopsy needles out. Instead, the goal is to fire as many needles that can hit the lesion - the learned strategy of the agent is to fire as close to the centre of the lesion as possible. 

\begin{figure}[ht!]

     \centering
     \begin{subfigure}[b]{\textwidth}
         \centering
         \includegraphics[width=\textwidth]{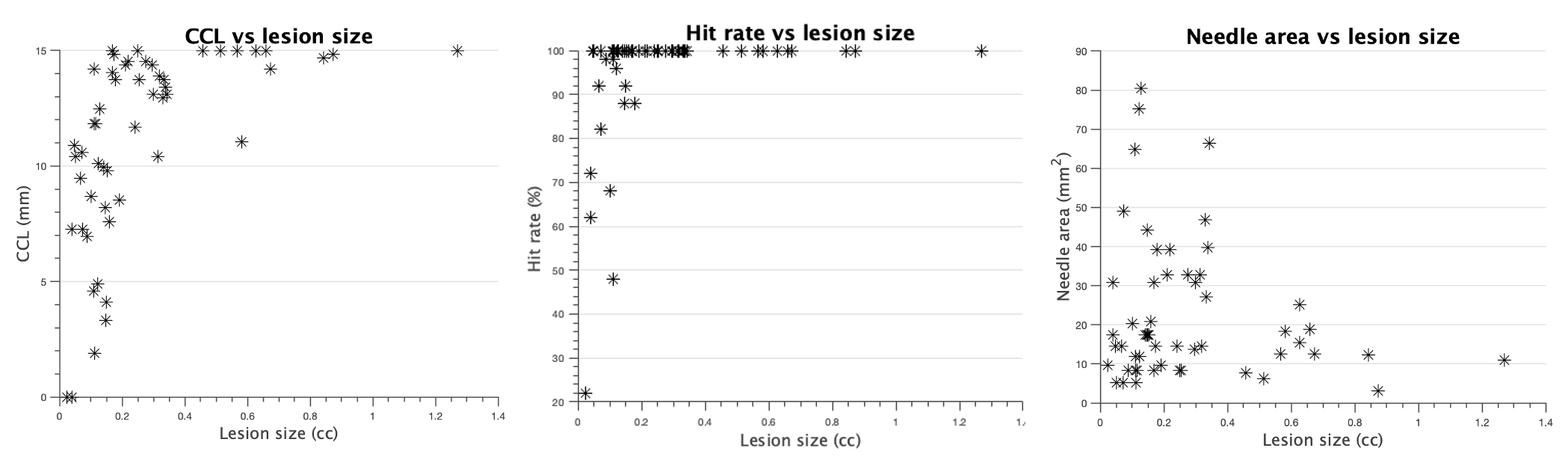}
    \caption{Biopsy metrics CCL, HR and NA vs lesion size} \label{fig:ccl_size}
     \end{subfigure}
    \hfill

    %\hfill
        
         \begin{subfigure}[b]{0.48\textwidth}
         \centering
         \includegraphics[width=\linewidth]{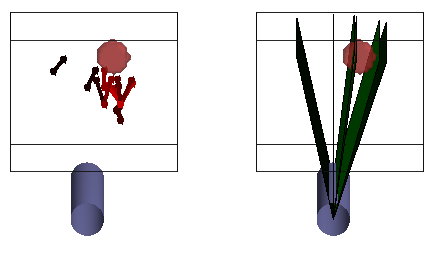}
         \caption{Small lesion size 0.2cc} 
         \label{fig:learned_policy_1}
     \end{subfigure}
     \begin{subfigure}[b]{0.48\textwidth}
         \centering
         \includegraphics[width=\textwidth]{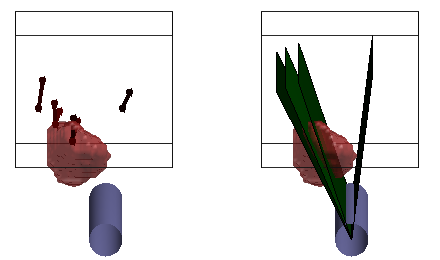}
         \caption{Large lesion size 0.4cc}
         \label{fig:policies}
     \end{subfigure}
        \caption{(a): CCL, HR and NA as a function of lesion size. (b) and (c): Examples of different sized targets (red), corresponding to the learned policies, represented by the needle sampling positions (red sticks, brighter indicates later time steps) and observed ultrasound images in green. The bounding cube and the cylinder represent the MR prostate volume and probe, respectively.}
        \label{fig:learned_plans}
\end{figure}
%\vspace{-5mm}
%2mm, 5mm

%%%\vspace{-2mm}
\vspace{-6pt}
\paragraph{Learned strategy for varying target sizes} - From Figure \ref{fig:ccl_size} and Table \ref{tab:small_vs_large}, we observe an interesting behaviour from the learned agent: 
%The measured CCL values increases, as lesion size increases, which shows that the obtained measurements are representative of the lesion size. For NA,  is observed : 
the smaller the lesions, the larger the spread of the needles. At a volume threshold of 0.4 cc, the mean CCL and NA are statistically different for smaller and larger lesions (p-values=0.002 \& 0.040), whilst the difference in HR is not (p-value=0.166). This result may seem counter-intuitive, as one would be cautious in spreading needles for a small target. However, the agent learned to distribute needles more widely for smaller lesions, attempting to maintain the hit rate, given the inevitable presence of target localisation uncertainty described in Sec.~\ref{sec:method.env}.
%This suggests that the learned strategies can effectively target lesions irrespective of their size, despite the differences in NA. 
Visual examples of the learned strategies are shown in Figures \ref{fig:learned_policy_1} and \ref{fig:policies}. 
%Since registration error was included in the simulation, this behaviour could have been learned in order to deal with this error. This increased coverage is important when the lesions are smaller and more difficult to target.  
This learned behaviour is interesting because a) it has not been observed previously, either in literature or in clinical practice. b) it improved the overall targeting performance compared to the target-size-agnostic baseline strategies and c) this could be suggested to urologists and interventional radiologists with or without the proposed RL assistance.

%In addition, an interesting observation is that the ultrasound image planes intersect with the lesion. Although not explicitly rewarded for this behaviour, the agent learns that finding image planes with the lesion is necessary to achieve a high reward.

\vspace{-5mm}
\begin{table}[ht!]
    \centering
    \caption{CCL, HR and NA for different lesion sizes using threshold size $<$ 0.4cc}
    \begin{tabular}{|c|c|c|c|}
    \hline
        \textbf{Lesion size} & \textbf{CCL $(mm)$ } & \textbf{HR $(\%)$ } &  \textbf{NA $(mm^2)$}    \\ 
         
         \hline
        Small lesions &  10.26 ± 4.19  & 93.16 ± 16.03  & 25.26 ± 19.45 \\ 
        \hline
        Large lesions & 14.52 ± 1.18 & 100.00 ± 0.00  & 13.02 ± 6.25   \\
        \hline
% & 42.85 ± 23.36
    \end{tabular}
    \label{tab:small_vs_large}
\end{table}

\section{Discussion and Conclusion}

The results show that the developed RL agents are competitive in sampling MR-derived targets, compared with intuitively devised strategies. Higher HR and average CCL were obtained by the agents, which was achieved by reducing the spread of the needles compared to baseline strategies. Furthermore, the learned strategies adapted to patient-specific procedures and varying pathology. The agents learned to achieve similar HR for different sized lesions, by spreading the fired needles more for smaller lesions. Such behaviour has not been observed before, and could be suggested to clinicians for improved targeting performance. Assumptions, such as number of allowed needles, template positioning and uncertainties in localisation/placement, have been made to facilitate the proposed pre-procedural planning. Some of them may be relaxed for an intra-procedural guidance tool - as a potential extension of this work, the others may require further validation. More importantly, the improved targeting performance provides means in mitigating the cancer under-sampling and help timely diagnosis of a significant number of prostate cancer patients with current MR-targeted biopsy.

\section*{Acknowledgement}
This work is supported by the EPSRC-funded UCL Centre for Doctoral Training in Intelligent, Integrated Imaging in Healthcare (i4health) [EP/S021930/1], EPSRC [EP/T029404/1], and the Department of Health’s NIHR funded Biomedical Research Centre at University College Hospital. This work is also supported by the International Alliance for Cancer Early Detection, an alliance between Cancer Research UK [C28070/A30912; C73666/A31378], Canary Center at Stanford University, the University of Cambridge, OHSU Knight Cancer Institute, University College London and the University of Manchester. This work was also supported by the Wellcome/EPSRC Centre for Interventional and Surgical Sciences [203145Z/16/Z].

\end{document}